\documentclass[runningheads]{llncs}
\usepackage{graphicx}

\usepackage{tikz}
\usepackage{comment}
\usepackage{amsmath,amssymb} 
\usepackage{color}

\usepackage[accsupp]{axessibility}  

\usepackage{booktabs}
\usepackage{cite}
\usepackage[colorlinks,citecolor=blue,urlcolor=magenta,bookmarks=false,hypertexnames=true]{hyperref} 

\newcommand*{\rev}{\textcolor{black}}
\newcommand*{\khan}{\textcolor{black}}

\begin{document}
\pagestyle{headings}
\mainmatter
\def\ECCVSubNumber{100}  
\title{LCDnet: A Lightweight Crowd Density Estimation Model for Real-time Video Surveillance} 

\titlerunning{ECCV-22 submission ID \ECCVSubNumber} 
\authorrunning{ECCV-22 submission ID \ECCVSubNumber} 
\author{Anonymous ECCV submission}
\institute{Paper ID \ECCVSubNumber}

\titlerunning{Lightweight Crowd Counting Models}
\author{Muhammad Asif Khan\inst{1} \and
Hamid Menouar \inst{1} \and
Ridha Hamila \inst{2}}  
\authorrunning{M. A. Khan et al.}
%
\institute{Qatar Mobility Innovations Center (QMIC), Qatar University \and
Electrical Engineering, Qatar University \\
\email{mkhan@qu.edu.qa},
\email{hamidm@qmic.com},
\email{hamila@qu.edu.qa}}
\maketitle

\begin{abstract}
Automatic crowd counting using density estimation has gained significant attention in computer vision research. As a result, a large number of crowd counting and density estimation models using convolution neural networks (CNN) have been published in the last few years. These models have achieved good accuracy over benchmark datasets. However, attempts to improve the accuracy often lead to higher complexity in these models. In real-time video surveillance applications using drones with limited computing resources, deep models incur intolerable higher inference delay. In this paper, we propose (i) a Lightweight Crowd Density estimation model (LCDnet) for real-time video surveillance, and (ii) an improved training method using curriculum learning (CL). LCDnet is trained using CL and evaluated over two benchmark datasets i.e., DroneRGBT and CARPK. Results are compared with existing crowd models. Our evaluation shows that the LCDnet achieves a reasonably good accuracy while significantly reducing the inference time and memory requirement and thus can be deployed over edge devices with very limited computing resources.
\dots
\end{abstract}

Crowd counting is an interesting research area that involves computer vision and deep learning to estimate the number of people in an images or video frames. Recently it has gain significant attention in the computer vision community due to the significance of the problem. Crowd counting is generally implemented in two ways: (i) counting objects (input is an image and the output is a number i.e., total head count in the image.), and (ii) density map estimation (input is an image and the output is the density map of the crowd which is then integrated to get the total head count.). Traditional methods for crowd counting were all based on detecting hand-crafted local features such as full body \cite{Topkaya_2014, Tuzel_2008}, body parts \cite{Li_2008, Viola_2001, Felzenszwalb_2010}, shapes \cite{Lin_2010}, or global features such as foreground \cite{Davies_1995}, edge \cite{Wu_2006}, texture \cite{Chen_2012} and gradient features \cite{Dalal_2005, Tian_2010} and then use machine learning models such as linear regression, ridge regression, Gaussian process, support vector machines (SVMs), random forest, gradient boost, and neural networks to provide the total count or a density map of the image. However, all the accuracy of these methods significantly degrades on images with dense crowds due to challenges such as occlusions, low resolution, foreshortening and perspectives.
\par
Recent research on crowd counting shows the efficacy of deep learning methods for crowd counting \cite{Khan2022RevisitingCC}. Convolution neural networks (CNNs) have a strong capability of auto feature extraction \cite{newR1, newR2} and perform very well in learning spatial features. Although even small CNN models \cite{CrowdCNN_CVPR2015} outperform traditional counting methods, their accuracy degrade on high density scenes.

To achieve higher accuracy in dense scenes, deeper models with a large size of parameters are developed \cite{CSRNet_CVPR2018, SANet_ECCV2018, CMTL_AVSS2017, MSFANet-ICPR2021}. These deep models although achieve good accuracy create performance bottlenecks in real-time applications due to large memory requirement, higher training complexity, and large inference delay. In contrast, small-sized CNN models offer several benefits in real-time video surveillance e.g., they incur low inference delay, require low memory for deployment on embedded devices, quickly update over-the-air, and can be trained, fine-tuned and run in distributed manner \cite{SqueezeNet_ICLR2017}. However, lightweight and shallow CNN models are usually disregarded by some due to their limited accuracy. Contrary to that, we believe that leveraging best practices such as carefully designing the model architecture, the use of accurately annotated training data, and efficient learning strategies can jointly improve the accuracy of shallow models to a greater extent \cite{Striving_for_simplicity_ICLR2015}. 

\subsection{Model Design Strategy}
Generally, a good choice of convolution filters play a crucial role in feature learning and contribute to reducing the size of the model. In essence, larger filters ($5\times5$) are more expensive and thus should be replaced by smaller filters ($3\times3$). As an example, a stack of three ($3\times3$) convolution layers is preferred over a single convolution layer of larger receptive fields such as $7\times7$ or $9\times9$ because of the non-linear activations between them. It also has less number of parameters ($3\times3^2\times C^2$ $<$ $1\times7^2\times C^2$ i.e., 81\% less parameters) and are computed faster. Further dimension reduction can be applied by using $a\times1$  convolution before expensive convolutions (e.g., $5\times5$ or $3\times3$) \cite{Inception_CVPR2015}. Furthermore, spatially separable convolutions are preferred i.e., a $3\times3$ convolution can be decomposed in two sequential $1\times3$ and $3\times1$ convolutions which leads to the same number of parameters and can even achieve better learning.

\subsection{Data Annotation Strategy}
In crowd density estimation, the point annotations on top of the heads in crowd image form a sparse "dot-map" or "localization-map" which is converted into a density map by convolving the head position with a Gaussian kernel. The scale parameter in the Gaussian kernel visually creates a blob around the point (pixel). A good density map is more accurate if the blob size covers the entire head and do not non-overlap with neighboring heads. However, due to the camera perspective effect, the size of heads vary in the same image. Adaptive kernels are used to select different values of the scale parameter which solve this problem to some extent. To cope with the perspective distortion and scale variations in crowd images, recent works propose often very deep models with complex architectures. However, unlike images captured with CCTV camera, in aerial images captured from drones, the perspective distortion is minimum and the scale variation is directly related to the drone altitude. If the drone-altitude is known, it can be used to generate accurate density maps.

\subsection{Model Learning Strategy}
The idea of curriculum learning (CL) in neural network was first presented in \cite{curriculum_learning_ICML2009}. The idea behind CL is the natural learning process in humans and animals i.e., humans learn better when concepts are presented in specific order of complexity i.e., from simpler to difficult tasks. Unlike traditional learning methods, in curriculum learning the training samples are sorted by order of (typically increasing) complexity. CL has been demonstrated as an effective strategy to improve the learning capability and faster convergence in various tasks e.g., computer vision \cite{Guy_2019, Guo_2018, Jiang_2014}, natural language processing (NLP) \cite{Platanios_2019, Tay_2019}, reinforcement learning \cite{Florensa_2017, Narvekar_2017, Ren_2018} etc. Recent studies \cite{Li_2021, DCL_2022} inspire to adopt CL in our research.

\rev{This work leverages the aforementioned three strategies for generating ground truth density maps and to design and train an extremely lightweight crowd density estimation model. In the first step, we designed the model (LCDnet) by carefully choosing convolution filters of different sizes in different layers. To keep the model more compact, we used less number of filters in the initial layers so even with the larger input feature maps, the computational load is controlled. To further alleviate the computational complexity, we used rectangular filters of size ($1\times3$) and ($3\times1$) instead of square filters. This also improves the learning performance as indicated in \cite{Inception_CVPR2015}. The resultant CNN model is a shallow network with only $0.05$ Million parameters. Next, to train the shallow model with drone-captured aerial images, we generated high quality (accurate) density maps. We tested different scale values of the Gaussian function and empirically found the most accurate values for each image. Lastly, the curriculum learning technique is employed to improve the learning performance of the model.}
\par
The contribution of our work is as follows:
\begin{itemize}
\item A lightweight CNN model (LCDnet) with fewer parameters, low memory requirement, and faster run-time than existing models. 
\item Generate density maps from sparse localization maps by considering drone-altitude to create adaptive Gaussian kernels to improve learning. 
\item Propose an efficient strategy based on curriculum-learning approach to further improve model training and convergence.
\item  Experimental demonstration of reasonably good performance using LCDnet on benchmark datasets. The accuracy is comparable to existing models of double-size than LCDnet.
\end{itemize}

\section{Related Work} \label {sec:rel_work}

A number of crowd counting datasets exists each of which can be divided into three categories. \textit{Surveillance-view} datasets containing indoor or outdoor images collected by surveillance cameras (e.g., Mall \cite{Mall_dataset2012}, UCSD \cite{UCSD_dataset2008}, WorldExpo'10 \cite{CrowdCNN_CVPR2015}, ShanghaiTech Part B \cite{MCNN_CVPR2016}, \textit{Free-view dataset} containing images from different sources including Internet (e.g., UCF-CC-50 \cite{UCF_CC_50_dataset2013}, UCF-QNRF \cite{CompositionLoss_2018}, ShanghaiTech Part A \cite{MCNN_CVPR2016}), and \textit{Drone view datasets} containing images collected using drones (e.g., DroneRGBT \cite{DroneRGBT_dataset}, CARPK \cite{CARPK_dataset}). Most of the earlier research works on crowd counting have been using the datasets in the first two categories. The drone-view datasets have been available recently.

While there has been different approaches for crowd counting, density estimation using CNN is the most widely used method. The first known CNN model for density estimation and counting is CrowdCNN \cite{CrowdCNN_CVPR2015}. The CrowdCNN model consists of three convolution layers followed by three fully connected layers. Following this, numerous works proposed different CNN-based models for crowd counting.

A multi-column CNN (MCNN) model is proposed in \cite{MCNN_CVPR2016} which consists of three CNN columns, each containing filters of receptive fields of different sizes (small, medium large). The outputs of the three columns are combined to predict the final density map. MCNN exhibits good performance to adapt to the scale variations in images due to perspective effects or different image resolutions.
A two column CNN model (CrowdNet) is proposed in \cite{CrowdNet_CVPR2016}. The model consists of two CNN columns i.e., a deep network (five CNN layers) and a shallow network (three CNN layers). The outputs of both networks are combined to predict the final density map.
A Switched CNN (SCNN) is proposed in \cite{SCNN_CVPR2017}, which consists of two parts, a Switch and a CNN regressor. The CNN regressor consists of three independent columns each with different size receptive fields. Patches from the input image are first fed to the Switch network, which relay it to one of the CNN regressor to predict the density map. The intuition in SCNN is to build CNN model which can adapt to the large scale variations without increasing the model computational complexity i.e., a patch is passed through only one column in the regressor. This reduces the computational complexity ascompared to other multi column CNN models e.g., MCNN and CrowdNet.
\par
Unlike multi-column CNN models, authors in \cite{MSCNN_ICIP2017} proposed a single column network  called multi scale CNN (MSCNN) to learn the scale variations. MSCNN uses three Inception modules \cite{Inception_CVPR2015} called multi-scale blobs (MSBs). Each MSB consists of multiple filters with different kernel size and is able to extract scale-relevant features.
The aforementioned CNN models can adapt to the scale variations introduced in the training data but may fail to generalize well \cite{CMTL_AVSS2017}. A cascaded multi-task learning (CMTL) model \cite{CMTL_AVSS2017} is proposed to adapt to the wide variations of density levels in images. CMTL is also a two column network. The first column is a high-level prior that classify an input image into groups based on the total count in the image. The features learned by the high level prior are shared with the second column that estimates the respective density map.
\par
The previous models mostly used multi-column architectures to learn scale-relevant features in crowd images achieves good results. To further improve the counting accuracy in highly congested scenes, authors in \cite{CSRNet_CVPR2018} propose deeper architecture by utilizing transfer learning. The Congested Scene Recognition (CSRNet) model \cite{CSRNet_CVPR2018} uses VGG16 \cite{VGG16_ICLR2015} (first 10 layers) as the front-end to extract features, and a back-end network with dilated convolution to substitute the pooling layers thus avoiding the loss of spatial information. Transfer learning an improve the feature extraction capability of the crowd counting model and has been recently adopted in CANNet \cite{CANNet_CVPR2019}, GSP \cite{GSP_CVPR2019}, TEDnet \cite{TEDnet_CVPR2019}, Deepcount \cite{DeepCount_ECAI2020}, SASNet \cite{SASNet_AAAI2021}, M-SFANet \cite{MSFANet-ICPR2021}, and SGANet \cite{SGANet_IEEEITS2022}.

\par
The aforementioned models often produces output density maps of lower resolution than the input image and uses patch-based training. The Trellis Encoder-Decoder Network (TEDnet) is proposed \cite{TEDnet_CVPR2019} which uses whole image as the model input and preserve the size of the density map to the actual resolution. Another encoder-decoder model with special Inception \cite{Inception_CVPR2015}-like modules is scale aggregation network (SANet) proposed in \cite{SANet_ECCV2018}. Like TEDnet, SANet also produces high resolution density map but has a simpler model structure.
\par

\rev{Our proposed model LCDnet simultaneously achieves two benefits as compared to the aforementioned models. First, it is extremely lightweight as compared to other models and can run faster even at edge devices with limited compute resources (a comparison is shown later in the paper). Secondly, it also produces density maps of better quality of size ($\frac{1}{2}$) of input size as compared to other methods e.g., CSRNet ($\frac{1}{8}$) and MCNN ($\frac{1}{4}$). It also requires least amount of memory to fit in on-chip caches.}

\vspace{-1em}

\section{Proposed Method} \label {sec:prop_method}
The aforementioned CNN-based crowd models are designed to improve counting accuracy. However, in a typical crowd monitoring system, the user may be interested in a rough estimation of crowd densities (e.g., low, medium, high, very high etc.) located in a geographical area rather than the exact count. Our goal in this paper is to develop a very lightweight model which can reasonably detect the presence of crowd scenes to an extent useful in analyzing crowds but essentially run faster on edge devices with limited computing resource. To this end, we propose a crowd density estimation model (LCDnet).


\subsection{Network Architecture}
\begin{figure*}[!h]
    \centering
    \includegraphics[width=0.98\textwidth]{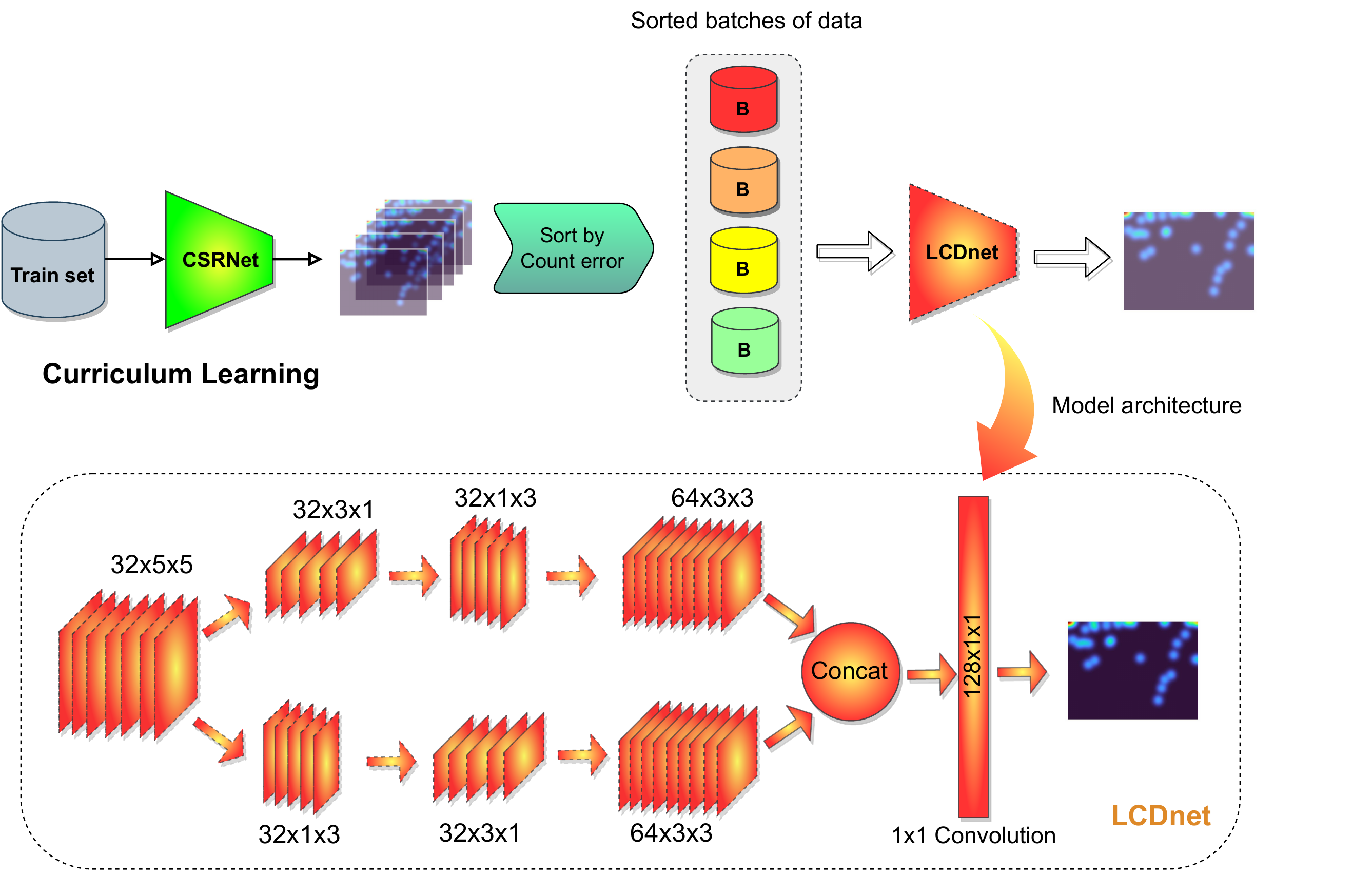}
    \caption{LCDnet with curriculum learning.}
    \label{fig:LCDnet}
\end{figure*}

The architecture of LCDnet is shown in Fig. \ref{fig:LCDnet}. It consists of six convolution (Conv) layers. Conv1 consists of 64 ($5\times5$) filters. The output of Conv1 layer is fed to Conv2 and Conv3, each having 32 ($3\times3$) filters. The outputs from Conv2 and Conv3 is fed to Conv4 and Conv5 both has 64 ($3\times3$)filters respectively. The outputs of Conv4 and Conv5 is concatenated and fed to Conv6 which consists of 128 ($1\times1$) filters to predict the final density map. It is worthy to note that LCDnet returns the density map rather than the crowd density class. There are two benefits for this. First, the density map preserves the location information of crowd which can be used to localize the crowd in the real-world. Second, it is easy to determine the total count (whole scene) or local count (specific part of the scene) from the predicted density map and even use the count information to configure user-defined crowd densities based on the application requirement.
\par
In the proposed LCDnet architecture, the first convolution layer is used to detect features such as edges. These detected features are then used in two columns. Both columns contain three layers i.e., two layers of 32 filters of size $1\times3$, and $3\times1$ (in reverse order in column 2), respectively, which is followed by a layer of 64 filters of $3\times3$ size. The outputs of both columns are concatenated and fed to a $1\times1$ conv layer of size 128, which generates a density map. The output density map is half size of the input image.

\subsection{Ground Truth Generation.}
If $x_i$ is a pixel containing the the head position, it can be represented by a delta function $\delta(x - x_i)$. The density map is generated by convolving the delta function with a Gaussian kernel $G_\sigma$.

\begin{equation}
    Y = \sum_{i=1}^{N}{ \delta(x-x_i) * G_\sigma}
\end{equation}
where, N is the total number of annotated points (i.e., total count of heads) in the image. The integral of density map $Y$ is equal to the total head count in the image. Visually, this operation creates a blurring of each head annotation using the scale parameter $\sigma$. There are various kernel settings to generate a variety of density maps.
The most basic approach is to keep $\sigma$ fixed value, which means that the density map will apply same kernel to all head positions irrespective of their scale in the image \cite{SANet_ECCV2018}.
As head sizes in image can vary due to camera prospective, a single value of $\sigma$ may not be a good choice. Hence, some recent works propose to use adaptive Gaussian kernels to create density maps. The value of $\sigma$ is calculated as the average distance to k-nearest neighboring head annotations. Visually, it generates lower degree of Gaussian blur for dense crowds and higher degree for region of sparse density in crowd scene. Typical settings includes $k=1$ \cite{CompositionLoss_2018}, $k=10$ \cite{SANet_ECCV2018, MSCNN_ICIP2017}.
Although adaptive Gaussian kernel may produce better results on images with large scale variations, our intuition is that drone images typically have less scale variations as compared to surveillance images e.g., from CCTV. The scale variations in drone images result from drone flying altitudes which do not vary too much due to regulatory measures. Thus, the scale variations are limited and a single value of $\sigma$ can be experimentally determined to produce density maps.
In our experiments, we empirically determined the value of $\sigma$ based on the drone altitudes. The datasets used in this study contain images captured from different altitudes. Thus, we first segregated images into different groups by drone altitudes and empirically found separate values of $\sigma$ for each group.

\subsection{Training.}
The image resolutions in the datasets used in this study are not very high, thus we use whole image-based training operations without extracting patches or downsampling operations on training images. However, to avoid model overfitting, data augmentations techniques such as horizontal flipping, and random brightness and contrast are applied. The kernels in all Conv layers are randomly initialized using Gaussian distribution with the value of standard deviations $0.01$. We used Adam optimizer with a base learning rate $0.0001$. The loss function used is pixel-wise euclidean distance between the target and predicted density maps which is defined in Eq. \ref{eq:mse}.

\begin{equation} \label{eq:mse}
    L(\Theta) = \frac{1}{N} \sum_{1}^{N}{ ||D(X_i;\Theta) - D_i^{gt}||_2^2}
\end{equation}

where $N$ is the number of samples in training data, $D(X_i;\Theta)$ is the predicted density map with parameters $\Theta$ for the input image $X_i$, and $D_i^{gt}$ is the ground truth density map.

We further applied curriculum learning technique to improve the learning performance of our model. In CL settings, we used transfer learning using CSRNet to determine the difficulty level of each image. Based on the counting error, images are sorted in ascending order before packing them into mini-batches. Thus, mini-batches are created in the order of their cumulative complexity.

\section{Experiments and Results} \label {sec:experiments}
The proposed model (LCDnet) was trained on a single GPU (Nvidia RTX-8000) using PyTorch deep learning framework. We also implemented and trained other models used in this study from the scratch for fair comparison.

\subsection{Datasets}
We evaluate the proposed model on two benchmark datasets i.e., DroneRGBT and CARPK. The DroneRGBT dataset contains images of people whereas the CARPK dataset contains images of cars, both captured from drones.
\par

\begin{figure}[!h]
\centering
\includegraphics[width=0.99\columnwidth]{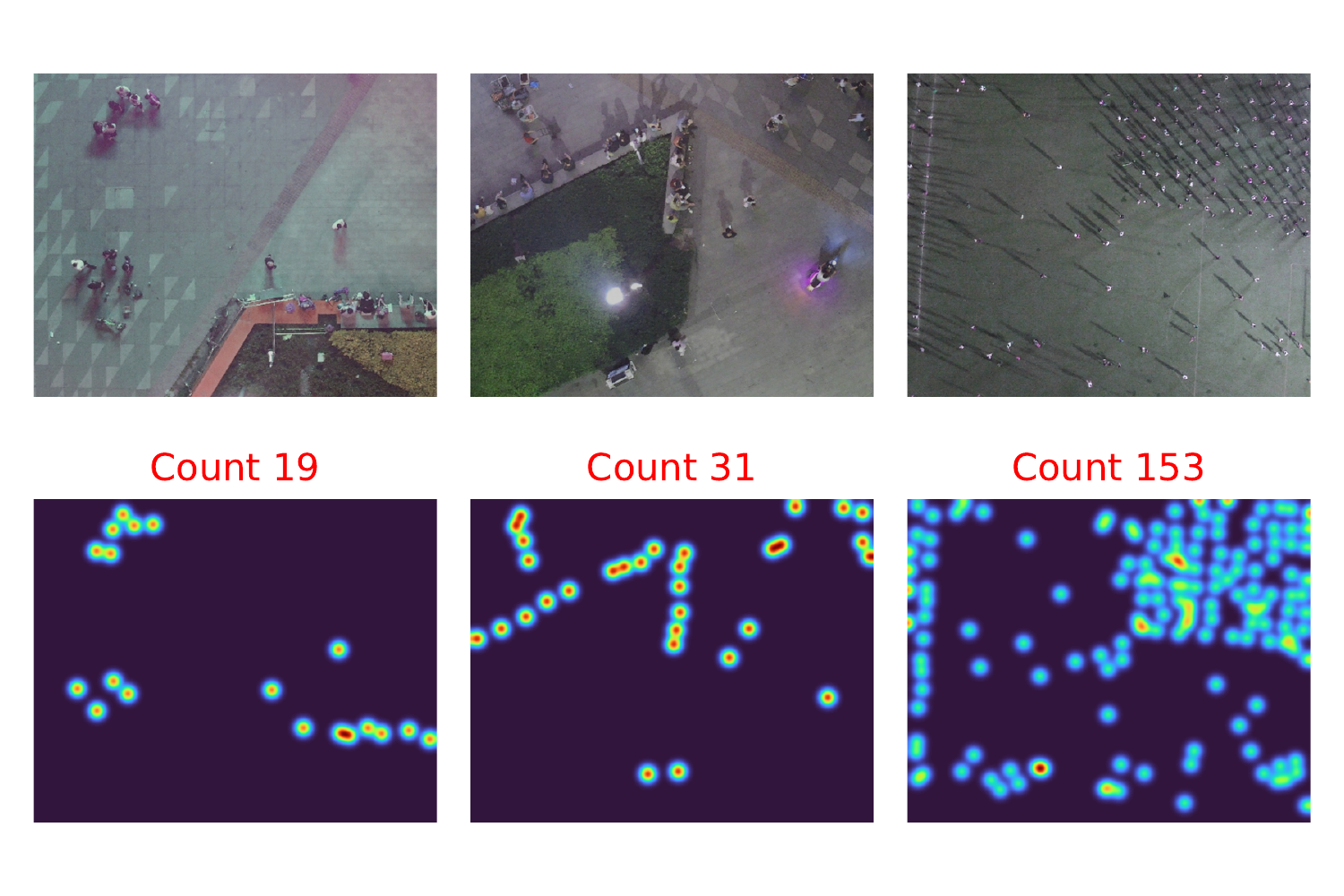}
\caption{Sample images (top) and their corresponding density maps (bottom) from DroneRGBT dataset.}
\label{fig:DroneRGBT_samples}
\end{figure}

\begin{figure}[!h]
\centering
\includegraphics[width=0.95\columnwidth]{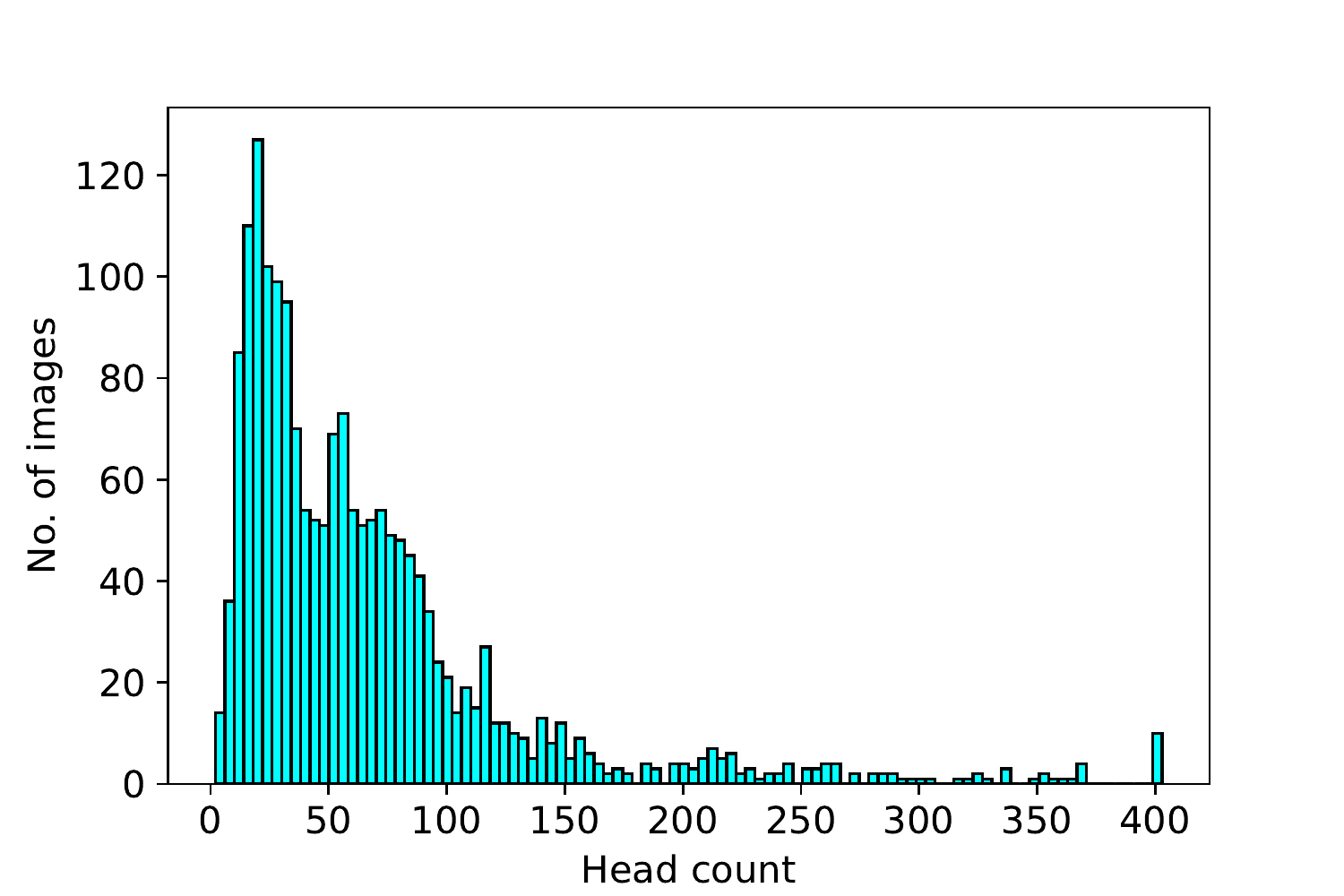}
\caption{Count distribution in DroneRGBT dataset.}
\label{fig:DroneRGBT_counts}
\end{figure}

\textbf{DroneRGBT.} The dataset contains $3600$ RGB and thermal image pairs with a spatial resolution of $512\times640$ pixels. The images cover a wide range of scenes e.g., campus, streets, parks, parking lots, playgrounds, and plazas. The dataset is divided into training set ($1807$ samples) and test set ($912$ samples) in such a way that both the training and test set include diverse images (i.e., different scenes, crowd densities, illumination, and scales) to avoid overfitting. The dataset provides head annotations of people. The count distribution and sample images from the dataset are presented in Fig. \ref{fig:DroneRGBT_counts} and Fig. \ref{fig:DroneRGBT_samples}, respectively.

\begin{figure}[!h]
\centering
\includegraphics[width=0.95\columnwidth]{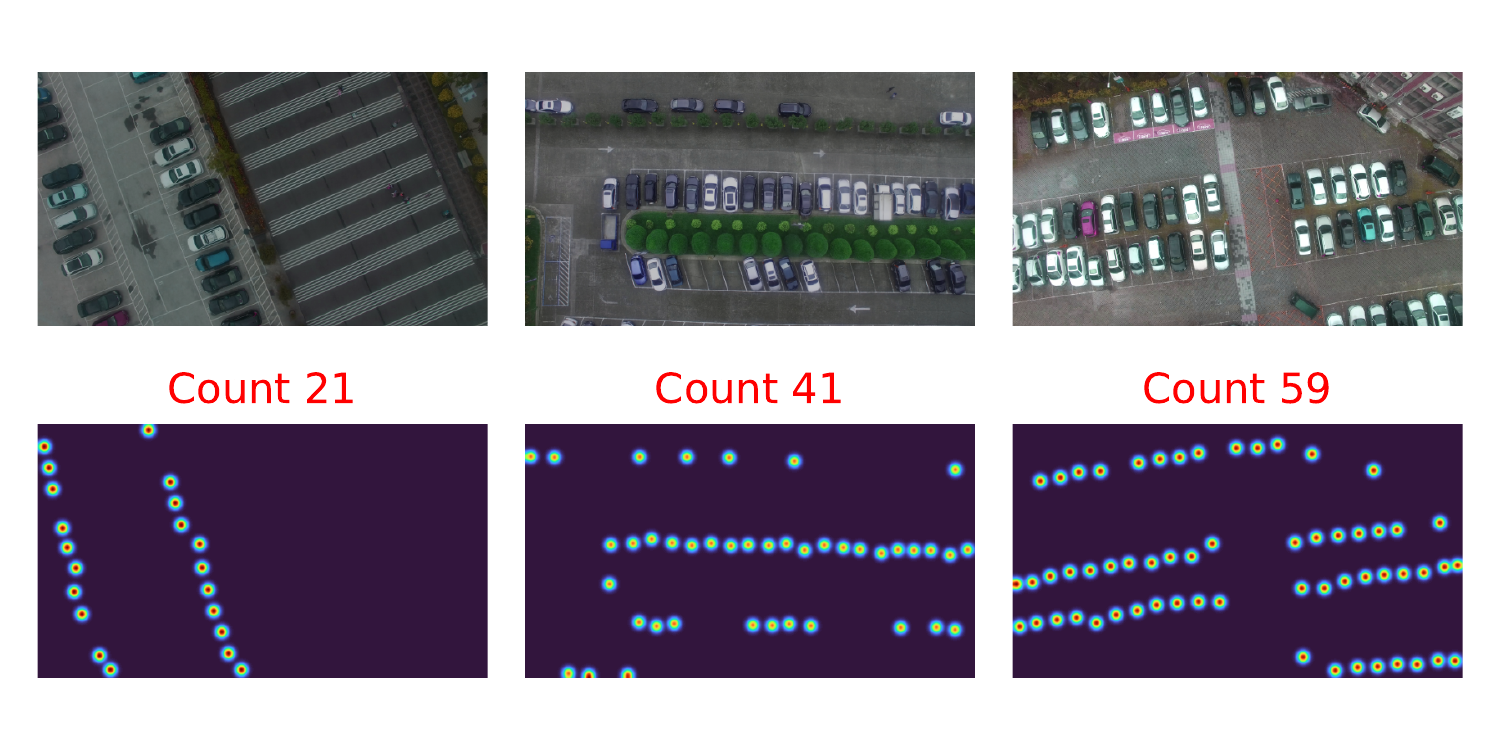}
\caption{Sample images (top) and their corresponding density maps (bottom) from CARPK dataset.}
\label{fig:CARPK_samples}
\end{figure}

\begin{figure}[!h]
\centering
\includegraphics[width=0.95\columnwidth]{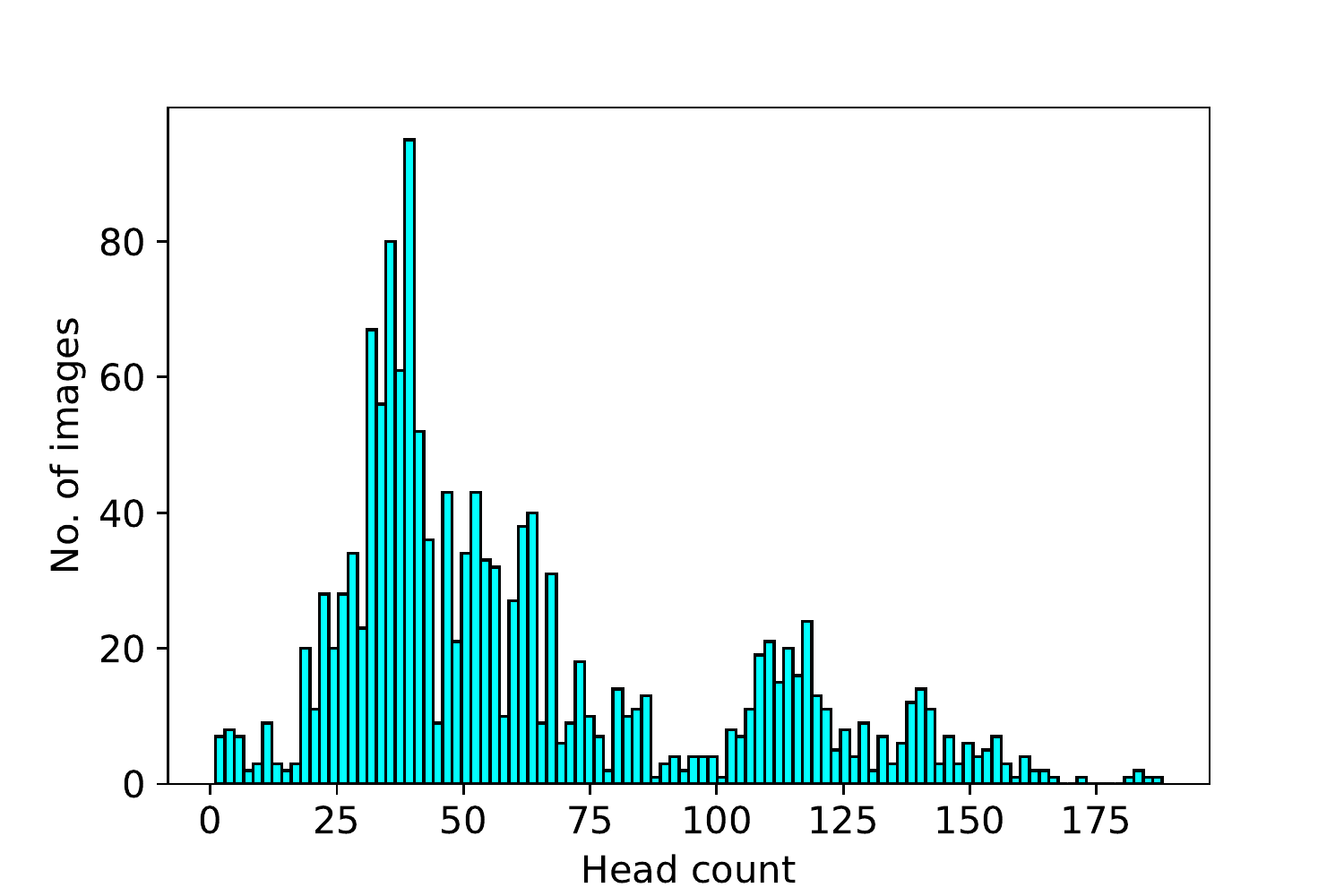}
\caption{Count distribution in CARPK dataset.}
\label{fig:CARPK_counts}
\end{figure}

\textbf{CARPK.} The dataset contains $1$ images of cars from from 4 different parking lots captured with a drone. The dataset is divided into a training set containing $989$ images and a test set containing $459$ images. The dataset has a total number of $89,777$ cars. The original dataset contains bounding box annotations, however we transformed the original annotations to dot annotation by taking the center of the bounding box. The count distribution and sample images from the dataset are presented in Fig. \ref{fig:CARPK_counts} and Fig. \ref{fig:CARPK_samples}, respectively.

\subsection{Evaluation Metrics}
Most of the existing works on crowd counting use mean absolute error (MAE) and Grid average mean absolute error (GAME) to evaluate the accuracy of the model. MAE is the average of absolute error in predicted counts and actual counts of all images. It is calculated using \ref{eq:mae}:

\begin{equation} \label{eq:mae}
    MAE = \frac{1}{N} \sum_{1}^{N}{(e_n - \hat{g_n})}
\end{equation}

where, $N$ is the total number of images in the dataset, $g_n$ is the ground truth (actual count) and $\hat{e_n}$ is the prediction (estimated count) in the $n^{th}$ image. While MAE is the most widely used metric in crowd counting research and is often used to compare various models, MAE provide image-wide counting and does not provide where the estimations have been done in the image. Owing to the possible estimation errors in MAE, authors in \cite{GAME_metric2015} proposed Grid Average Mean absolute Error (GAME). In GAME, an image is divided into $4^L$ non-overlapping patches and compute MAE separately for each patch. Thus, GAME poses a more robust and accurate estimation for crowd counting applications. It is defined in q. \ref{eq:game}.

\begin{equation} \label{eq:game}
    GAME = \frac{1}{N} \sum_{n=1}^{N}{ ( \sum_{l=1}^{4^L}{|e_n^l - g_n^l|)}}
\end{equation}

The GAME metric is more robust to localisation errors in density estimation by calculating localized error among the target and predicted density maps. We set the value of $L=4$, thus each density map is divided into a grid size of $4\times4$ creating $16$ patches. The absolute difference in the head count for each patch is measured and summed for all patches of the same density map, then averaged over the whole dataset. We compared the LCDnet model against existing models over the two metrics to provide a fair evaluation of the model accuracy. However, the true benefit of LCDnet is the lower model complexity at the cost of tolerable counting error. In addition, the performance is measured over two other metrics i.e., structural similarity index (SSIM), peak signal-to-noise ratio (PSNR). Both SSIM and PSNR evaluate the quality of the predicted density maps and are measured in Eq. \ref{eq:ssim} and \ref{eq:psnr} as follows:

\begin{equation} \label{eq:ssim}
    SSIM (x,y) = \frac{(2\mu_x \mu_y + C_1)  (2\sigma_x \sigma_y C_2)}  {(\mu_z^2 \mu_y^2 + C_1)  (\mu_z^2 \mu_y^2 + C_2)}
\end{equation}
where $\mu_x, \mu_y, \sigma_x, \sigma_y$ represents the means and standard deviations of the actual and predicted density maps, respectively.

\begin{equation} \label{eq:psnr}
    PSNR = 10 log_{10}\left( \frac{Max(I^2)}{MSE}  \right)
\end{equation}
where $Max(I^2)$ the maximal in the image data. If it is an 8-bit unsigned integer data type, the $Max(I^2)=255$.

\subsection{Evaluation Results}
We compared the proposed model (LCDnet) against two mainstream crowd density estimation models i.e., MCNN \cite{MCNN_CVPR2016}, and CSRNet \cite{CSRNet_CVPR2018}. MCNN is a relatively small-sized CNN model which has gained good counting accuracy over several benchmark datasets as compared to other models of similar size. CSRNet on the other hand is a deep CNN model that uses VGG-16 \cite{VGG16_ICLR2015} as a front-end and has shown high accuracy in dense crowd scenes. Both models have been used for comparison in many crowd counting studies and thus we chose them as candidate models for small-sized and large-sized models in this study. \par

We compare the performance of LCDnet in terms of counting accuracy at the cost of model size and complexity against the MCNN and CSRNet on DroneRGBT and CARPK datasets. While the primay comparison is done against MCNN and CSRNet, we additionally provide complexity comparison against some other well-known counting models to highlight the benefit of the proposed model (LCDnet). The model complexity comparison is shown in Table \ref{tab:results_complexity} whereas the accuracy comparison is depicted in Table \ref{tab:results_error}.
The inference time is computed on GPU server (Nvidia RTX 8GB), and two different edge devices (Nvidia Jetson Xavier and Jetson Nano). The system details of these devices are as follows:
\begin{itemize}
\item \textbf{Server:} GPU (Nvidia Quadro RTX-8000).
\item \textbf{Jetson Xavior NX:} 64bit system with Processor (6-core NVIDIA Carmel ARM), Memory (8GB), GPU (NVIDIA Volta architecture with 384 NVIDIA CUDA cores and 48 Tensor cores).
\item \textbf{Jetson Nano:} 64bit system with Quad-Core Arm Cortex-A57 MPCore, Memory (4GB), GPU (128-core NVIDIA Maxwell GPU).
\end{itemize}


\begin{table*}[!h]
\centering
\renewcommand*{\arraystretch}{1.5}
\caption{Comparison of \rev{proposed scheme (LCDnet trained with curriculum learning) against SOTA models} for number of parameters (in Million), GMACs, size (in MB), and inference time (in mili seconds) for fixed input size.}
\label{tab:results_complexity}
\begin{tabular}{p{2.5cm} ccccccc} \toprule
Model    &Output &Parameters (M)  &Size (MB)   &GMACs &\multicolumn{3}{c}{Inference Time (s)}  \\ \midrule 

& & & & &Server &Jetson Xavier &Jetson Nano \\ \midrule
CrowdCNN \cite{CrowdCNN_CVPR2015}    &1/4 &1.66 &6.65  &3.96 &0.071 &0.073 &0.23\\ 
MCNN \cite{MCNN_CVPR2016}    & 1/4 &0.13 &0.53 &8.82   &0.05 &0.10 &0.21\\
CMTL  \cite{CMTL_AVSS2017}   &1/4 &2.45  &9.82 &39.82 &0.098 &0.31 &0.62\\ 
CSRNet \cite{CSRNet_CVPR2018}  &1/8   &16.26 &65.05  &135.4 &0.19 &1.01 &1.88\\
SANet \cite{SANet_ECCV2018}   &1/4 &0.25 &1.02  &8.97 &0.075 &0.12 &0.23\\
\textbf{LCDnet (ours)}   &\textbf{1/2} &\textbf{0.05}  &\textbf{0.21} &\textbf{4.85} &\textbf{0.006} &\textbf{0.05} &\textbf{0.10}\\ \bottomrule
\end{tabular}
\end{table*}

\begin{table*}[!h]
\centering
\caption{Accuracy comparison of the \rev{proposed scheme (LCDnet trained with curriculum learning) against SOTA models} over DroneRGBT dataset \cite{DroneRGBT_dataset} and CARPK dataset \cite{CARPK_dataset}.}
\label{tab:results_error}
\begin{tabular}{cccccccccc} \toprule

Method  &\multicolumn{4}{c}{DroneRGBT}  & &\multicolumn{4}{c}{CARPK} \\ \cmidrule{2-5}  \cmidrule{7-10}
&MAE &GAME &SSIM &PSNR & &MAE &GAME &SSIM &PSNR\\ \midrule \midrule

\rev{CrowdCNN \cite{CrowdCNN_CVPR2015}} &\rev{26.6} &\rev{48.2} &\rev{0.52} &\rev{17.3} & &\rev{15.6} &\rev{49.1} &\rev{0.63}  &\rev{18.8} \\[0.2em]

MCNN \cite{MCNN_CVPR2016}            &17.9  &44.49 &0.54 &18.2 &   &10.3 &42.40 &0.76 &19.21 \\[0.2em]

\rev{
CMTL \cite{CMTL_AVSS2017}}            &\rev{18.1} &\rev{40.5} &\rev{0.53} &\rev{17.1} &   &\rev{10.2} &\rev{41.3} &\rev{0.73} &\rev{19.0} \\[0.2em]

CSRNet \cite{CSRNet_CVPR2018}        &7.6 &25.7 &0.72 &21.70 &   &6.12 &21.8 &0.82 &20.52\\[0.2em]

\khan{SANet \cite{SANet_ECCV2018}} &\rev{16.2} &\rev{34.5} &\rev{0.59} &\rev{19.4} & &\rev{9.8} &\rev{27.2} &\rev{0.76}  &\rev{19.8} \\[0.2em]

\textbf{LCDnet (ours)}       &\textbf{21.4} &\textbf{46.9} &\textbf{0.60} &\textbf{21.39} &   &\textbf{13.1} &\textbf{45.2} &\textbf{0.79} &\textbf{20.14} \\[0.2em]

\bottomrule
\end{tabular}
\end{table*}

On DroneRGBT dataset, LCDnet achieves MAE of $21.4$ which is comparable with that of MCNN ($17.9$). In terms of complexity, LCDnet has almost half the number of parameters and half number of multiply-add-calculations (GMACs). LCDnet also incurs much lower ($\frac{1}{2} \times$) inference delay as compared to MCNN. 
Although the accuracy of CSRNet is much higher than both LCDnet ($3\times$) and MCNN ($2.2\times$), it has a very huge size requiring large memory size and much higher ($20\times$) inference delay than LCDnet.

\begin{figure*}[!h]
\centering
\includegraphics[width=0.99\textwidth]{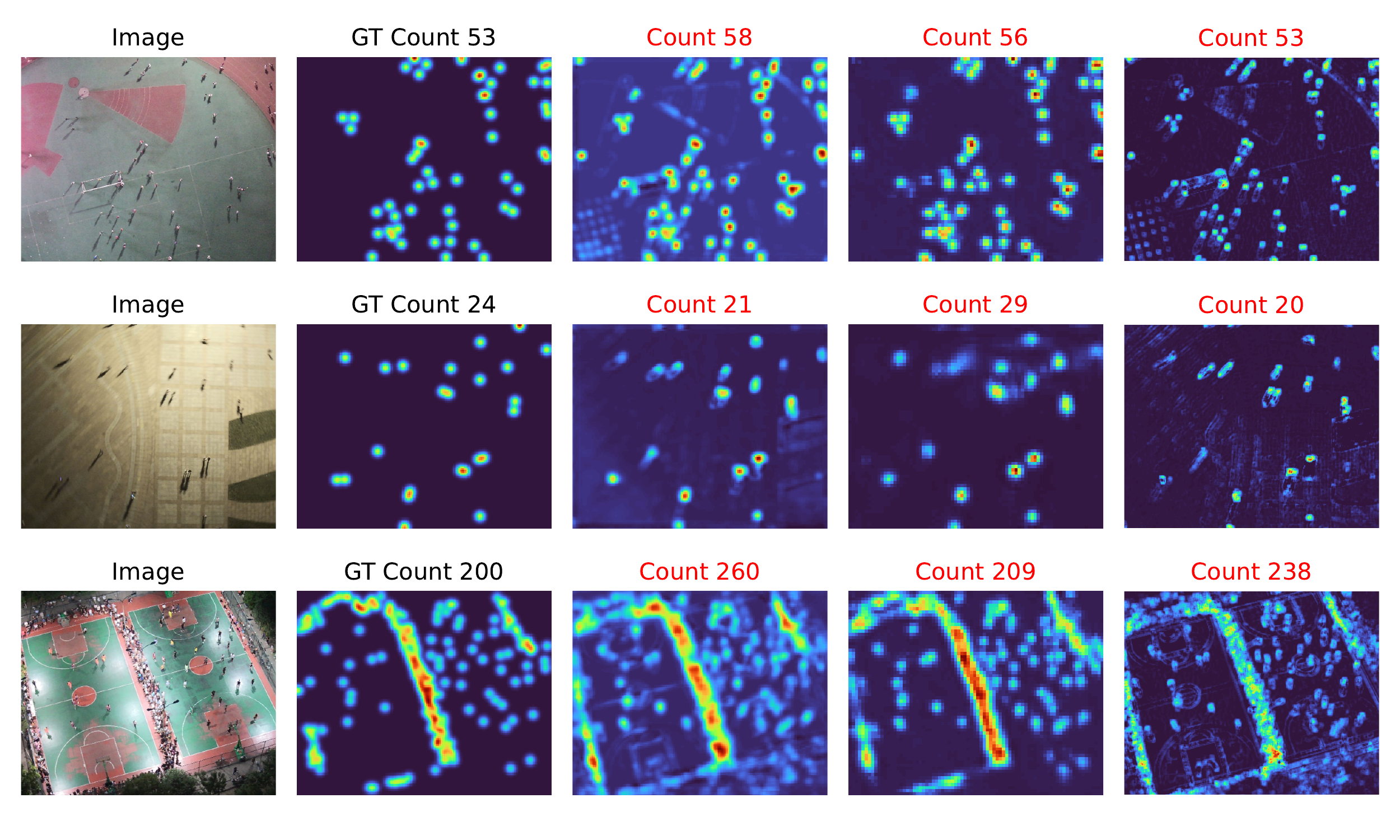}
\caption{Comparison of predictions on on DroneRGBT dataset. The first two columns shows crowd images and their corresponding ground truth. Columns 3-4 shows predictions using MCNN \cite{MCNN_CVPR2016} and CSRNet \cite{CSRNet_CVPR2018} without curriculum learning. Column 5 shows predictions using LCDnet (ours), respectively.}
\label{fig:DroneRGBT_compare}
\end{figure*}

\begin{figure*}[!h]
\centering
\includegraphics[width=0.99\textwidth, height=7cm]{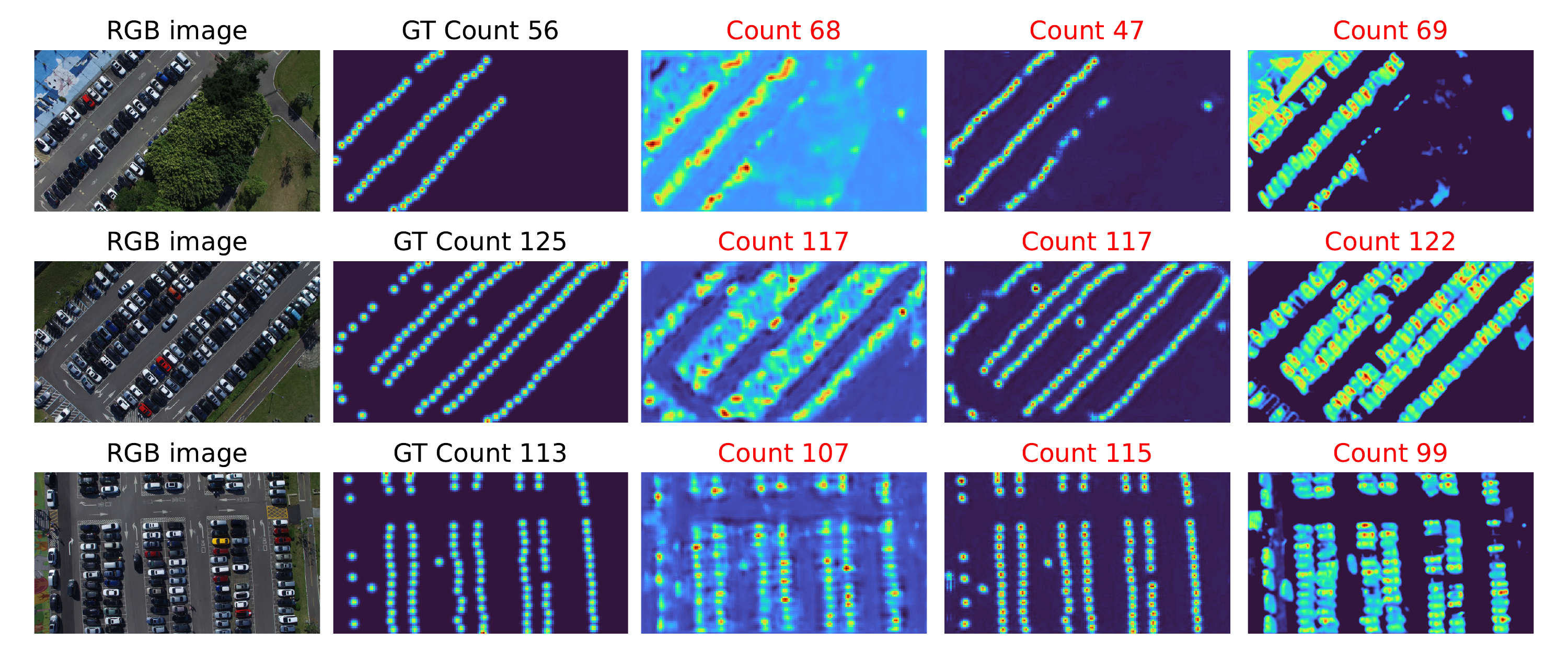}
\caption{Comparison of predictions on on CARPK dataset. The first two columns shows crowd images and their corresponding ground truth. Columns 3-4 shows predictions using MCNN \cite{MCNN_CVPR2016} and CSRNet \cite{CSRNet_CVPR2018}. without curriculum learning. Column 5 shows predictions using LCDnet (ours), respectively.}
\label{fig:CRAPK_compare}
\end{figure*}

On CARPK dataset, LCDnet achieves better results. It achieves MAE $13.1$, which is close to MCNN ($10.1$) and slighltly less than CSRNet ($6.12$). Some sample predictions using MCNN, CSRNet and the proposed LCDnet models over DroneRGBT and CARPK datasets, respectively, are shown in Fig. \ref{fig:DroneRGBT_compare} and Fig. \ref{fig:CRAPK_compare}. It can be visualized that LCDnet has good detection capability and produces better quality density maps than MCNN for DroneRGBT dataset. We believe this is due to the use of small sized filters ($1\times3$ and $3\times1$). The better quality of density map is expected and evident from the higher values of SSIM and PSNR.

\section{Conclusion} \label{sec:conclusion}

This paper proposes a lightweight crowd density estimation model (LCDnet) for deployment over resource-constrained embedded devices (e.g., drones) suitable for real-time applications (e.g., surveillance) scenarios. The paper outlines various design principles and best practices used to develop efficient CNN architectures. LCDnet is designed by adopting three efficient strategies; (i) compact CNN model (ii) improved ground truth generation from head annotations and drone altitudes, and (iii) improved training mechanism using curriculum learning. LCDnet is evaluated on two different datasets of drone-captured images i.e., DroneRGBT, CARPK. Our experimental analysis shows that LCDnet achieves reasonably good accuracy at much lower computational cost. The small memory footprint and lower inference time makes LCDnet a good fit for drone-based video surveillance.

\clearpage
%
%
\bibliographystyle{splncs04}
\bibliography{egbib}
\end{document}